\documentclass[10pt, conference]{IEEEtran}
\IEEEoverridecommandlockouts
\usepackage{cite}
\usepackage{amsmath,amssymb,amsfonts}
\usepackage{graphicx}
\usepackage{textcomp}
\usepackage{xcolor}
\usepackage{hyperref}
\usepackage{algorithm}
\usepackage{algorithmicx}
\usepackage{algpseudocode}
\usepackage{booktabs}
\usepackage{balance}
\def\BibTeX{{\rm B\kern-.05em{\sc i\kern-.025em b}\kern-.08em
    T\kern-.1667em\lower.7ex\hbox{E}\kern-.125emX}}

\makeatletter
\newcommand\fs@betterruled{%
  \def\@fs@cfont{\bfseries}\let\@fs@capt\floatc@ruled
  \def\@fs@pre{\vspace*{5pt}\hrule height.8pt depth0pt \kern2pt}%
  \def\@fs@post{\kern2pt\hrule\relax}%
  \def\@fs@mid{\kern2pt\hrule\kern2pt}%
  \let\@fs@iftopcapt\iftrue}
\floatstyle{betterruled}
\restylefloat{algorithm}
\makeatother

\makeatletter
\renewcommand{\fnum@algorithm}{\textbf{Algorithm~\thealgorithm}}
\makeatother

\begin{document}

\title{3D Dynamic Radio Map Prediction Using Vision Transformers for Low-Altitude Wireless Networks}

\author{
\IEEEauthorblockN{Nguyen Duc Minh Quang\IEEEauthorrefmark{1}\IEEEauthorrefmark{2}, Chang Liu\IEEEauthorrefmark{1}, Huy-Trung Nguyen\IEEEauthorrefmark{2}, Shuangyang Li\IEEEauthorrefmark{3},\\ Derrick Wing Kwan Ng\IEEEauthorrefmark{4}, and Wei Xiang\IEEEauthorrefmark{1}}
\IEEEauthorblockA{\IEEEauthorrefmark{1}School of Computing, Engineering, and Mathematical Sciences, La Trobe University, Australia}
\IEEEauthorblockA{\IEEEauthorrefmark{2}Data Governance Laboratory, Posts and Telecommunication Institute of Technology, Vietnam}
\IEEEauthorblockA{\IEEEauthorrefmark{3}Chair of Communications and Information Theory, Technical University of Berlin, Germany}
\IEEEauthorblockA{\IEEEauthorrefmark{4}School of Electrical Engineering and Telecommunications, University of New South Wales, Australia}
Email: \{quang.nguyen, c.liu6, w.xiang\}@latrobe.edu.au, trungnh@ptit.edu.vn,\\ shuangyang.li@tu-berlin.de, w.k.ng@unsw.edu.au
}

\IEEEaftertitletext{\vspace{-1\baselineskip}}
\maketitle

\begin{abstract}
Low-altitude wireless networks (LAWN) are rapidly expanding with the growing deployment of unmanned aerial vehicles (UAVs) for logistics, surveillance, and emergency response. Reliable connectivity remains a critical yet challenging task due to three-dimensional (3D) mobility, time-varying user density, and limited power budgets. The transmit power of base stations (BSs) fluctuates dynamically according to user locations and traffic demands, leading to a highly non-stationary 3D radio environment. Radio maps (RMs) have emerged as an effective means to characterize spatial power distributions and support radio-aware network optimization. However, most existing works construct static or offline RMs, overlooking real-time power variations and spatio-temporal dependencies in multi-UAV networks. To overcome this limitation, we propose a {3D dynamic radio map (3D-DRM)} framework that learns and predicts the spatio-temporal evolution of received power. Specially, a Vision Transformer (ViT) encoder extracts high-dimensional spatial representations from 3D RMs, while a Transformer-based module models sequential dependencies to predict future power distributions. Experiments unveil that 3D-DRM accurately captures fast-varying power dynamics and substantially outperforms baseline models in both RM reconstruction and short-term prediction.
\end{abstract}

\begin{IEEEkeywords}
radio map, low-altitude wireless network, UAV communication, power prediction, spatio-temporal transformer
\end{IEEEkeywords}

\section{Introduction}

Low-altitude wireless networks (LAWN) leverage the sub-1000-m airspace to support large-scale UAV operations for various applications such as logistics, surveillance, and disaster response~\cite{jiang_integrated_2025, zeng_accessing_2019}. 
However, ensuring reliable communication in these highly dynamic environments remains a major challenge due to rapid three-dimensional (3D) mobility, frequent handovers, and time-varying propagation conditions~\cite{zeng_cellular-connected_2019}. 
Furthermore, base stations (BSs) are required to dynamically adapt their transmit power according to the spatial distribution of aerial users, causing the received power field to fluctuate across both space and time. 
As such, accurate modeling and prediction of these variations are critical for ensuring link stability and efficient resource allocation. 
In this regard, radio maps (RMs) have emerged as an effective approach to represent spatial radio characteristics~\cite{romero2022radio}, thereby enabling location-aware UAV trajectory planning and adaptive power control~\cite{zhang_radio_2021, zhang2024fast}. 
These factors motivate the need for precise \emph{3D dynamic RMs} to enhance situational awareness and communication reliability in LAWNs.

Recent studies have investigated deep-learning-based RM construction, with growing interest in generative artificial intelligence (AI) models~\cite{nguyen2024deep, liu2020deepresidual, liu2020deeptransfer, lxm2020deepresidual} such as generative adversarial networks (GANs) and diffusion models~\cite{zhang2024fast, zhang2023rme, wang2024radiodiff, zhao20253d, hu20233d}, which have demonstrated superior reconstruction accuracy compared with traditional methods such as Kriging and spatial interpolation~\cite{sato2017kriging}. However, most existing approaches are designed for static RM reconstruction, failing to capture real-time evolution of received power field. For instance, Zhang {et al.}~\cite{zhang2024fast} proposed a conditional GAN-based framework for fast 2D RM estimation with reduced data requirements, however similar to~\cite{zhang2023rme, wang2024radiodiff}, it remains limited to two-dimensional (2D) spatial representations, which cannot fully characterize 3D LAWNs. Although these methods reduce computational complexity to a certain extent, they lack temporal modeling capability. In contrast, Zhao {et al.}~\cite{zhao20253d} introduced a diffusion model-based 3D RM achieving high accuracy and sampling efficiency. Nevertheless, the iterative nature of diffusion models results in high inference latency, limiting their suitability for real-time prediction. More importantly, most prior works focus on static maps and overlook the dynamic evolution of received power, an essential aspect for LAWN scenarios with mobile UAVs and adaptive BS power allocation.

To overcome these limitations, we develop a {3D dynamic radio map (3D-DRM)} framework for real-time power prediction in LAWNs. The proposed framework jointly reconstructs and forecasts 3D RMs through an end-to-end Transformer architecture, where a Vision Transformer (ViT)~\cite{dosovitskiy2020image} encoder learns spatial features and a temporal Transformer module~\cite{vaswani2017attention} models sequential dependencies for short-term prediction. This unified design captures both spatial correlations and temporal dynamics, enabling strong generalization across diverse UAV densities, trajectories, and network configurations. To the best of our knowledge, this is the first work to address {3D dynamic RM prediction} for UAV-enabled LAWN employing a ViT-based architecture for unified spatio-temporal learning. The main contributions are summarized as follows:
\begin{itemize}
\item We formulate the problem of {3D dynamic RM prediction}, addressing two inherently coupled challenges: reconstructing the 3D RM from sparse UAV measurements and forecasting its short-term evolution under dynamic BS allocation and UAV mobility.
\item We propose the {3D-DRM framework}, which jointly solves both challenges through a unified Transformer architecture integrating a ViT-based spatial encoder and a temporal Transformer predictor.
\item We validate the proposed 3D-DRM under diverse network conditions, including different UAVs densities and mobility patterns, demonstrating its robustness and clear improvements over baseline models.
\end{itemize}

\vspace{-2pt}
\section{System Model and Problem Formulation}
\label{sec:system-model}

\begin{figure}
    \centering
    \includegraphics[width=1\linewidth]{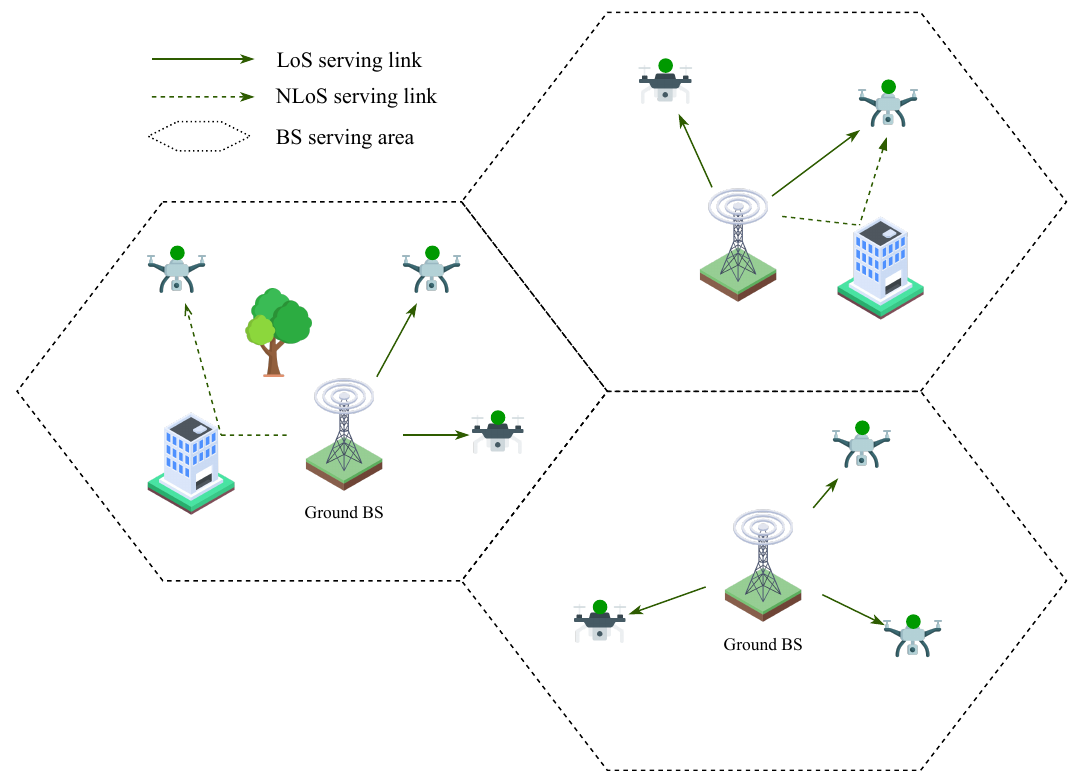}\vspace{-10pt}
    \caption{An illustration of the cellular-connected low-altitude wireless network.}\vspace{-5pt}
    \label{fig:system-model}
\end{figure}

In this work, we consider a cellular-connected LAWN consisting of $M$ ground base stations (GBSs) and $Q$ UAVs, as illustrated in Fig.~\ref{fig:system-model}. Let $\mathbf{p}_q(t)\!\in\!\mathbb{R}^3$ denote the 3D position of UAV $q, \forall q\in\{1,\dots,Q\}$ at time $t$. 
Each GBS $m, \forall m \in \{1,\dots, M\}$ is equipped with $N_t$ transmit antennas and a total downlink power budget ${P}_m$, serving a subset of UAVs within its coverage area, each equipped with $N_r$ receive antennas.

\subsection{Power Allocation and Beamforming Model}
Each GBS performs downlink multi-user transmission, that simultaneously serves a subset of UAVs, denoted by $\mathcal{S}_m(t)$, at time $t$, with cardinality $|\mathcal{S}_m(t)|\!\le\!N_t$. 
The UAV mobility model is characterized by~\cite{liu2022learning}
\begin{equation}
\mathbf{p}_q(t+\Delta_t)
= \mathbf{p}_q(t) + \mathbf{v}_q(t)\,\Delta_t,
\label{eq:mobility_model}
\end{equation}
where $\mathbf{v}_q(t)\!\in\!\mathbb{R}^3$ denotes its instantaneous velocity vector and $\Delta_t$ is the sampling interval between consecutive time steps. 

All GBSs are assumed to employ an {equal power allocation (EPA)}~\cite{castaneda2016overview, yuan2020learning, liu2022deep} policy across their active data streams\footnote{In practice, two UAVs rarely occupy the same angular direction relative to a GBS due to their spatial separation and mobility. Hence, potential beam overlap or interference between co-directional users is assumed negligible in this work~\cite{krishnan2017spatio}.}.
Specifically, let $d_{m,q}(t)\!=\!\|\mathbf{p}_q(t)-\mathbf{l}_m\|_2$ denote the distance between GBS $m$ and UAV $q$, where $\mathbf{l}_m$ is the known GBS location. 
The power allocated from GBS $m$ to UAV $q\!\in\!\mathcal{S}_m(t)$ is defined as
\begin{equation}
{P}_{m, q}(t)=
\frac{d_{m,q}^\alpha(t)}{\sum_{q'\in\mathcal{S}_m(t)} d_{m,q'}^\alpha(t)}\,{P}_m,
\label{eq:distance_weighted_power}
\end{equation}
where $\alpha> 0$ is a tunable distance-weighting exponent that controls the compensation strength for far users.
Let $\mathbf{v}_{m,q}(t)\!\in\!\mathbb{C}^{N_t\times1}$ denote the precoding vector of GBS $m$ for UAV $q\!\in\!\mathcal{S}_m(t)$, and $s_{m,q}(t)$ the transmitted symbol with $\mathbb{E}[|s_{m,q}(t)|^2]\!=\!1$. 
The transmitted signal from GBS $m$ is then expressed as
\begin{equation}
\mathbf{x}_m(t)=
\sum_{q\in\mathcal{S}_m(t)}
\sqrt{{P}_{m, q}(t)}\,\mathbf{v}_{m,q}(t)\,s_{m,q}(t),
\label{eq:tx_signal}
\end{equation}
which satisfies $\mathbb{E}[\|\mathbf{x}_m(t)\|^2]={P}_m$. 
Each UAV employs a receive steering vector $\mathbf{a}_q(t) \in \mathbb{C}^{N_r\times1}$ that aligns toward the dominant direction of its serving GBS to maximize received power. The corresponding effective channel from GBS $m$ to UAV $q$ is defined as
\begin{equation}
\mathbf{h}_{m,q}(t)=\mathbf{H}_{m,q}^H(t)\,\mathbf{a}_q(t),
\label{eq:effective_channel}
\end{equation}
where $\mathbf{H}_{m,q}(t)\!\in\!\mathbb{C}^{N_r\times N_t}$ denotes the MIMO channel matrix.

\subsection{Power Measurement Model and Dynamic Radio Map}
The total received power at UAV $q$ and time $t$ is given by
\begin{equation}
y_q(t)=\sum_{m=1}^{M}\sum_{k\in\mathcal{S}_m(t)}
{P}_{m, q}(t)
\big|\mathbf{h}_{m,q}^H(t)\mathbf{v}_{m,k}(t)\big|^2 + w_q,
\label{eq:rssi_user}
\end{equation}
where $w_q$ denotes receiver noise power.  Beside, the summation captures the total received power from all GBSs, where the beamforming gains $\big|\mathbf{h}_{m,q}^H\mathbf{v}_{m,k}\big|^2$ are dominant for the serving beam~\cite{liu2023predictive, liu2022learning, liu2020location}.

The 3D environment is discretized into uniform voxels of size $\Delta_d$, forming a spatial grid $\mathcal{V}\!\in\!\mathbb{R}^{W\times L\times H}$. 
Within each voxel, the propagation conditions and large-scale power variations are assumed approximately constant, allowing all measurements within the same cell to represent a single spatial sample. 
Time is similarly discretized with a step $\Delta_t$, corresponding to a short-term averaging window adopted to smooth temporal fluctuations. 
During each interval, the UAV displacement is negligible relative to $\Delta_d$, and the set of scheduled users at each GBS is assumed to vary slowly~\cite{krishnan2017spatio}. Accordingly, the averaged measurement at UAV $q$ is given by
\begin{equation}
\begin{aligned}
    \bar{y}_q(t_n)&=\frac{1}{\Delta_t}\!\int_{t_n-\Delta_t}^{t_n}\!y_q(\tau)\,d\tau \\
   &\approx\!\sum_{m=1}^{M}\bar{g}_m(\bar{\mathbf{p}}_q)\,\bar{P}_m(t_n;\Delta_t)+w_q,
\end{aligned}
\label{eq:avg_rssi_user}
\end{equation}
where $\bar{\mathbf{p}}_q\!\in\!\mathcal{V}$ denotes the voxel containing UAV $q$, 
$\bar{g}_m(\mathbf{p})=\mathbb{E}_{k}[|\mathbf{h}_{m,q}^H\mathbf{v}_{m,k}|^2]$ represents the average channel gain, 
and $\bar{P}_m(t_n;\Delta_t)$ is the average transmit power of GBS $m$ over $\Delta_t$. 
Here, $t_n=n\Delta_t$ denotes the discretized time instant. 
Because variations in $\bar{P}_m(t_n;\Delta_t)$ within a short window are negligible compared with spatial power differences across voxels~\cite{zhao20253d}, it can be approximated as $\bar{P}_m(t_n;\Delta_t)\!\approx\!\bar{P}_m(t_n)$. The dynamic RM is then defined as
\begin{equation}
\mathcal{P}(\mathbf{p},t_n)=\sum_{m=1}^{M}\bar{g}_m(\mathbf{p})\,\bar{{P}}_m(t_n),
\label{eq:power_map_user}
\end{equation}
which characterizes the spatial distribution of the averaged received power at voxel $\mathbf{p}$ and sampling time $t_n$.

\subsection{Problem Formulation}
\label{sec:problem-formulation}

This work aims to predict the short-term dynamic RM $\mathcal{P}(t_n)$ defined in~\eqref{eq:power_map_user}. Specifically, given historical power measurements collected by UAVs along their trajectories at prior sampling instants, the objective is to estimate the upcoming 3D RM over the considered area. Let $\mathcal{Y}_{1:n}=\{(\bar{y}_q(t_i),\mathbf{p}_q(t_i))\}_{i=1}^{n}$ denote the set of estimated received power measurements and the corresponding UAV locations up to time $t_n$, conventional RM estimation works~\cite{zhang2024fast, zhang2023rme, wang2024radiodiff, zhao20253d, hu20233d} focus on reconstructing a static RM by minimizing the discrepancy between the ground-truth RM $\mathcal{P}$ and an estimated map $\hat{\mathcal{P}} = R_\omega(\mathcal{Y})$, where $R_\omega(\cdot)$ is an estimator parameterized by $\omega$. However, our goal is to predict the short-term sequence of RMs at future times $t_{n+1}$ to $t_{n+T}$. This is a coupled task, as it must reconstruct the current RM and predict its short-term evolution.
Accordingly, the short-term RM prediction can be formulated as
\begin{equation}
\begin{aligned}
    &\min_{F_{\Theta}}~ 
    \mathbb{E}_{\bar{\mathcal{Y}}_{n+1:n+T} \,|\, \mathcal{Y}_{1:n}}
    \bigg\{
    \sum_{\tau=1}^{T}
    \sum_{\mathbf{p} \in \mathcal{V}}
    \big\|
    \mathcal{P}(\mathbf{p},t_{n+\tau}) 
    - 
    \hat{\mathcal{P}}(\mathbf{p},t_{n+\tau})
    \big\|
    \bigg\},\\
    &\text{s.t.}~ \eqref{eq:tx_signal},~\eqref{eq:avg_rssi_user}, \quad \hat{\mathcal{P}}(t_{n+1:n+T}) = F_\Theta(\mathcal{Y}_{1:n}),
\end{aligned}
\end{equation}
where $F_\theta$ is the mapping function parameterized by $\Theta$, and $\mathbb{E}_{\bar{\mathcal{Y}}_{n+1:n+T} \,|\, \mathcal{Y}_{1:n}}\{\cdot\}$ denotes the conditional expectation over future measurement sequences $\bar{\mathcal{Y}}_{n+1:n+T}$ (i.e., the future sets of received-power observations), given the historical measurements $\mathcal{Y}_{1:n}$. Since only historical measurements are available when reconstructing the RM and predicting its short-term evolution, the conditional ergodic average is adopted to characterize the performance of reconstruction and prediction~\cite{liu2022learning}.

Inferring a complete 3D RM from partial observations and forecasting its short-term evolution under dynamic channel and mobility conditions are inherently coupled tasks. Accurate prediction requires coherent spatial representations, whereas reliable reconstruction depends on accurately modeling temporal dynamics. Model-based methods are intractable due to the need for explicit joint spatio–temporal channel models, while existing data-driven approaches focus on static reconstruction rather than temporal evolution. Hence, an effective framework must jointly learn spatial completion and temporal prediction in a unified manner~\cite{liu2022scalable}.

\section{3D-DRM: 3D Dynamic Radio Map Predictor}
\label{sec:methodology}

To address the aforementioned problem, we propose the 3D-DRM, which learns a unified spatio-temporal representation of the RM leveraging an encoder–decoder Transformer architecture.  The following subsections describe the preliminaries of Transformers, the overall architecture, its encoder and decoder designs, and the learning algorithm employed for training and inference.

\begin{figure}[t]
    \centering
    \includegraphics[width=0.9\linewidth]{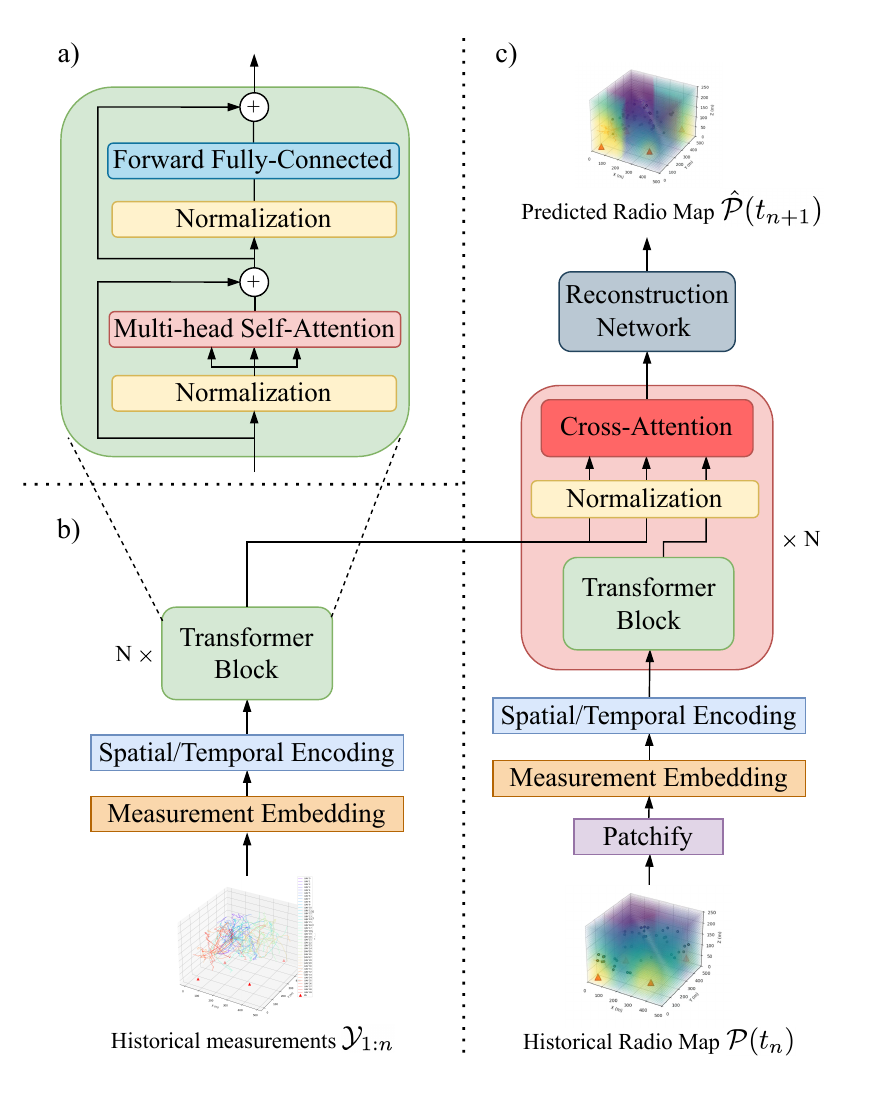}\vspace{-5pt}
    \caption{Overall architecture of the proposed 3D-DRM framework. 
    (a) Transformer block used in both the encoder and decoder to model token-wise attention; 
    (b) The encoder embeds irregular UAV measurements into latent features through spatial attention, while the temporal module captures their dynamic evolution over time; 
    (c) The decoder applies cross-attention between voxel queries and latent embeddings to reconstruct the complete 3D RM.}\vspace{-10pt}
    \label{fig:3d_dpm_framework}
\end{figure}

\subsection{Preliminaries of Transformer Models}
\label{sec:preliminaries}

This work builds on the Transformer architecture~\cite{vaswani2017attention}, which effectively captures dependencies in sequential and spatial data. A Transformer models a mapping $\mathbf{X}\!\rightarrow\!\hat{\mathbf{Y}}$, where the input tokens $\mathbf{X}=\{\mathbf{x}_1,\dots,\mathbf{x}_N\}$ represent local information units (e.g., power measurements at spatial--temporal coordinates), and the output tokens $\hat{\mathbf{Y}}=\{\hat{\mathbf{y}}_1,\dots,\hat{\mathbf{y}}_M\}$ represent reconstructed or predicted elements. The model comprises an encoder $f_\theta$ that generates contextual embeddings $\mathbf{H}_e=f_\theta(\mathbf{X})$ and a decoder $d_\phi$ that produces outputs conditioned on these embeddings, $\hat{\mathbf{Y}}=d_\phi(\mathbf{H}_e)$. 
Central to this framework is the attention mechanism~\cite{vaswani2017attention}, which computes pairwise token relevance via
$a_{ij}=\mathrm{softmax}\!\left(\tfrac{Q_i K_j^{\top}}{\sqrt{d_k}}\right),$
where $Q_i=\mathbf{x}_i W_Q$ and $K_j=\mathbf{x}_j W_K$ are learned projections of $\mathbf{x}_i$ and $\mathbf{x}_j$, with $W_Q, W_K \in \mathbb{R}^{d_{\text{model}}\times d_k}$, $d_k$ and $d_{model}$ denoting head dimension and embedding dimension of the tokens, respectively. This operation enables each token to aggregate global context without predefined spatial or temporal neighborhoods. In fact, by stacking attention and feed-forward layers, the encoder captures long-range spatio-temporal dependencies, while the decoder refines predictions through cross-attention. 
In our design, 3D-DRM employs a measurement encoder that embeds irregular UAV observations into latent tokens and a power-map decoder that reconstructs the full 3D RM via cross-attention, as illustrated in Fig.~\ref{fig:3d_dpm_framework}.

\subsection{Power Measurement Encoder}
\label{subsec:encoder}

The encoder is designed to process the unordered and spatially irregular set of UAV power measurements denoted by
$\mathcal{Y}_{1:n}=\{(\bar{y}_q(t_i),\mathbf{p}_q(t_i))\}_{i=1}^{n}$.
Each measurement is first embedded into a high-dimensional token representation:
\begin{equation}
\mathbf{z}_{q,i} = (W_e\,\bar{y}_q(t_i) + b_e) + \phi(\mathbf{p}_q(t_i)) + \tau(t_i),
\label{eq:encoder_token}
\end{equation}
where $W_e \in \mathbb{R}^{d_{\text{model}}}$ and $b_e$ are learnable embedding parameters, 
$\phi(\mathbf{p}_q(t_i))$ is the spatial positional encoding mapping 3D coordinates to sinusoidal Fourier features, 
and $\tau(t_i)$ denotes the temporal encoding capturing the sampling time~\cite{vaswani2017attention}.  

Let $\mathbf{H}_e^{(0)} = [\mathbf{z}_{1,1}, \mathbf{z}_{1,2}, \dots, \mathbf{z}_{Q,n}]$ denotes the matrix of embedded measurement tokens.  
The encoder is composed of $L_e$ stacked Transformer blocks, 
each containing a multi-head self-attention (MSA) sublayer, a feed-forward network (FFN), 
and residual connections followed by layer normalization (LN).  
At the $\ell$-th layer, the latent features are updated as~\cite{dosovitskiy2020image}
\begin{equation}
\begin{aligned}
\tilde{\mathbf{H}}_e^{(\ell)} 
&= \mathbf{H}_e^{(\ell-1)} + \text{MSA}\!\left( 
\text{LN}\big(\mathbf{H}_e^{(\ell-1)}\big)\right), \\[4pt]
\mathbf{H}_e^{(\ell)} 
&= \tilde{\mathbf{H}}_e^{(\ell)} + \text{FFN}\!\left( 
\text{LN}\big(\tilde{\mathbf{H}}_e^{(\ell)}\big)\right),
\end{aligned}
\label{eq:encoder_layer}
\end{equation}
where $\tilde{\mathbf{H}}_e^{(\ell)}$ represents the intermediate features after attention aggregation, 
and $\mathbf{H}_e^{(\ell)}$ denotes the output features of the $\ell$-th encoder layer after nonlinear transformation.  
Here, $\text{MSA}(\cdot)$ captures pairwise spatial dependencies among tokens, 
while $\text{FFN}(\cdot)$ models higher-order feature interactions.  
Through this mechanism, the encoder learns spatial attention patterns reflecting propagation similarities among measurement locations, 
implicitly modeling large-scale fading and interference characteristics.  
The final output $\mathbf{H}_e^{(L_e)}$ represents the latent measurement embeddings passed to the temporal prediction module.

\subsection{Radio Map Decoder}
\label{subsec:decoder}

The decoder reconstructs the complete 3D RM at the next time step $t_{n+1}$ 
based on the latent measurement representations extracted by the encoder. 
Let the voxel grid $\mathcal{V}\subset\mathbb{R}^{W\times L\times H}$ be partitioned into
non-overlapping cubic patches of side $P$ voxels (voxel size $\Delta_d$), yielding
a patch index set $\mathcal{R}=\{1,\dots,R\}$ and disjoint voxel sets
$\{\,\mathcal{B}_r\,\}_{r\in\mathcal{R}}$ with $|\mathcal{B}_r|=P^3$ and
$\bigcup_{r}\mathcal{B}_r=\mathcal{V}$~\cite{dosovitskiy2020image}. 
To incorporate both spatial-temporal context and local power information,
each 3D patch $\mathcal{B}_r(t_n)$ extracted from the latest observed RM 
$\mathcal{P}(t_n)$ is flattened and linearly embedded as
\begin{equation}
\mathbf{e}_r(t_n) = W_p\,\text{vec}\big(\mathcal{B}_r(t_n)\big) + b_p,
\end{equation}
where $\text{vec}(\cdot)$ flattens all voxel powers within patch $r$,
and $W_p,b_p$ are learnable patch embedding parameters.  
Each embedded patch is then augmented with spatial and temporal encodings to form a query token:
\begin{equation}
\mathbf{q}_r(t_{n+1}) = 
\mathbf{e}_r(t_n)
+ \phi(\mathbf{p}_r)
+ \tau(t_{n}),
\label{eq:decoder_query_corrected}
\end{equation}
where $\phi(\mathbf{p}_r)$ encodes the 3D position of patch $r$ and $\tau(t_{n})$ provides the temporal cue for the prediction step. 
The resulting set $\mathbf{Q}(t_{n+1})=\{\mathbf{q}_r(t_{n+1})\}_{r=1}^R$  is further passed through Transformer blocks as $\mathbf{H}_e$ in Eq.~\eqref{eq:encoder_layer}, then serves as the query input to the decoder.

To inject historical information, cross-attention is performed between the decoder queries $\mathbf{Q}(t_{n+1})$ and the encoded latent context $\mathbf{H}_e^{(L_e)}$~\cite{vaswani2017attention}, yielding the context-aware patch embeddings $\tilde{\mathbf{Q}}=\{\tilde{\mathbf{q}}_r\}_{r=1}^R$. These attended vectors are intermediate latent representations that integrate both spatial structure and temporal context.  
A reconstruction head $d_{\phi}$ then maps each attended patch feature to per-voxel power values:
\begin{equation}
\hat{\mathcal{P}}_r(t_{n+1})=d_{\phi}(\tilde{\mathbf{q}}_r).
\label{eq:decoder_output_final}
\end{equation}
The predicted patches $\{\hat{\mathcal{P}}_r(t_{n+1})\}_{r=1}^{R}$ are assembled to form the complete predicted RM $\hat{\mathcal{P}}(t_{n+1})$.
During training, the ground-truth map $\mathcal{P}(t_n)$ is available and used as input to guide the learning process. 
During inference, when RMs are unavailable, the decoder is initialized with a start token representing $\mathcal{P}(t_n)$ and performs autoregressive prediction, i.e., recursively feeding its own previous output $\hat{\mathcal{P}}(t_{n+i-1})$ as input to generate $\hat{\mathcal{P}}(t_{n+i})$ for $i\!\in\![1,k]$.

\subsection{Learning Algorithm of 3D-DRM}
\label{subsec:learning}

\begin{algorithm}[tp]
\caption{\strut\strut\strut\strut Training and Inference Procedure of 3D-DRM}
\label{alg:training-inferring}
\begin{algorithmic}
\Require Measurements $\mathcal{Y}_{1:n}$, GBS locations $\mathbf{L}$, ground-truth maps $\mathcal{P}(t_{1:n+k})$
\Ensure Predicted maps $\hat{\mathcal{P}}(t_{n+1:n+k})$
\vspace{5pt}
\State \textbf{Training:}
\While{not converged}
    \State Encode $\mathcal{Y}_{1:n}$ using $L_e$ Transformer blocks $\Rightarrow \mathbf{H}_e^{(L_e)}$
    \For{$i=1$ to $k$}
        \State Embed patches from $\mathcal{P}(t_{n+i-1})$ with spatial \& temporal encoding
        \State Perform cross-attention: $\tilde{\mathbf{Q}} \!\leftarrow\! \mathrm{CrossAttn}(\mathbf{Q}, \mathbf{H}_e^{(L_e)})$
        \State Reconstruct $\hat{\mathcal{P}}(t_{n+i}) = d_\phi(\tilde{\mathbf{Q}})$
    \EndFor
    \State Compute total loss $\mathcal{L}_{\text{total}}$; update parameters
\EndWhile
\State \textbf{Inference:}
    \State Encode $\mathcal{Y}_{1:n}$ to obtain $\mathbf{H}_e^{(L_e)}$
    \State Initialize $\tilde{\mathcal{P}}(t_n)$ with a start token
    \For{$i=1$ to $k$}
        \State Form $\mathbf{Q}$ from $\tilde{\mathcal{P}}(t_{n+i-1})$
        \State $\tilde{\mathbf{Q}} \!\leftarrow\! \mathrm{CrossAttn}(\mathbf{Q}, \mathbf{H}_e^{(L_e)})$
        \State $\hat{\mathcal{P}}(t_{n+i}) = d_\phi(\tilde{\mathbf{Q}})$
        \State $\tilde{\mathcal{P}}(t_{n+i}) \!\leftarrow\! \hat{\mathcal{P}}(t_{n+i})$
    \EndFor
\end{algorithmic}
\end{algorithm}

The 3D-DRM model is trained to minimize discrepancies between predicted and ground-truth RMs while maintaining temporal consistency. 
The total loss is formulated as a weighted sum of two complementary terms,
\begin{equation}
\mathcal{L}_{\text{total}} =
\gamma\,\mathcal{L}_{\text{vox}}
+ \beta\,\mathcal{L}_{\text{temp}},
\label{eq:loss_total}
\end{equation}
where $\gamma$ and $\beta$ balance the contributions of voxel-wise accuracy and temporal smoothness, respectively.
Especially, the voxel-wise fidelity loss~\cite{romero2022radio} $\mathcal{L}_{\text{vox}}$ enforces numerical accuracy between the predicted and reference RMs at each voxel, defined as
\begin{equation}
\mathcal{L}_{\text{vox}} =
    \sum_{\tau=1}^{k}
    \sum_{\mathbf{p} \in \mathcal{V}}
    \big\|
    \mathcal{P}(\mathbf{p},t_{n+\tau}) 
    - 
    \hat{\mathcal{P}}(\mathbf{p},t_{n+\tau})
    \big\|_2^2.
\label{eq:loss_vox}
\end{equation}
Additionally, to promote temporal coherence and suppress abrupt changes across consecutive time steps, a temporal-gradient loss~\cite{xiao2022learning} is applied:
\begin{equation}
\mathcal{L}_{\text{temp}} =
\sum_{\tau=1}^{H}
\big\|
\nabla_t \hat{\mathcal{P}}(t_{n+\tau})
- \nabla_t \mathcal{P}(t_{n+\tau})
\big\|_2^2,
\label{eq:loss_temp}
\end{equation}
where $\nabla_t$ denotes the first-order temporal difference operator. 
This encourages the smooth evolution of the power distribution over time while preserving the motion dynamics of the RM.
By jointly optimizing these three objectives, 3D-DRM learns to generate RMs that are accurate and temporally stable. A summary of both training and inferring procedures is presented in Algorithm~\ref{alg:training-inferring}.

\section{Experiment Results}
In the following section, we first describe the simulation setup, baselines, and metrics, and the performance evaluation among various settings.

\subsection{Simulation Setup}

\subsubsection{System Settings}
The configuration of the data generation process is summarized in Table~\ref{tab:data_config}. 
In each simulated scene, a given number of BSs is first deployed within the service area. 
Multiple UAVs are then initialized with randomly assigned start and destination points and navigate between them along randomized trajectories. 
Along these trajectories, the UAVs collect averaged received signal strength measurements, which are used as inputs to the prediction model.
To ensure diversity and broad coverage of operating conditions, the dataset is generated under varying network configurations, including different numbers of BSs and UAVs, and distinct UAV trajectories. This setup produces heterogeneous radio environments for robust model training and evaluation.

\begin{table}[t]
\centering
\caption{Simulation Configuration}
\begin{tabular}{l l}
\toprule
\textbf{Parameter} & \textbf{Value / Range} \\
\midrule
Map size & $500\times500\times250~\mathrm{m}^3$ \\
Voxel resolution & 2\,m \\
BS count & 4 \\
UAV count & 40--80 \\
UAV velocity & 5--10\,m/s \\
Transmit power & 30--45\,dBm \\
Noise figure & 3--7\,dB \\
Bandwidth & 100\,MHz \\
\bottomrule
\end{tabular}\vspace{-10pt}
\label{tab:data_config}
\end{table}

\subsubsection{Training and Evaluation Settings}
All simulations are executed on a workstation equipped with 2$\times$AMD EPYC 7313 CPUs, 514\,GiB RAM, and 4$\times$NVIDIA RTX 3090 GPUs (48\,GB VRAM each).
The synthetic dataset consists of 120 sequences, each containing 200 frames, where UAVs move dynamically across the frame and collect measurements along their trajectories, and BSs are randomly distributed.
For each sequence, the first 150 frames are exploited for model training, and the remaining 50 frames are used for testing. 
The model input is the set of UAV measurement tokens collected over the past 15 frames, and the task is to predict the next 5 frames of the 3D RM.

\subsubsection{Benchmarks and Metrics}
For fair comparison, two baselines are considered: RadioUNet~\cite{levie2021radiounet}, the benchmark for RM reconstruction, and ConvLSTM~\cite{huang2021spatial}, a convolutional recurrent model suited for spatio-temporal forecasting.
Note that RadioUNet is adapted to predict future RMs rather than only reconstructing the current map.
We report two metrics aligned with the learning objectives:
root mean squared error (RMSE)~\cite{zhao20253d} measures voxel-wise reconstruction fidelity;
temporal gradient error quantifies temporal smoothness and motion consistency across consecutive frames.

\subsection{Evaluation Results}

\begin{figure*}[t]
\centering
\renewcommand{\arraystretch}{1.0}
\setlength{\tabcolsep}{-2pt} 

\begin{tabular}{c c c}
\raisebox{-0.5\height}{%
    \includegraphics[width=0.35\textwidth]{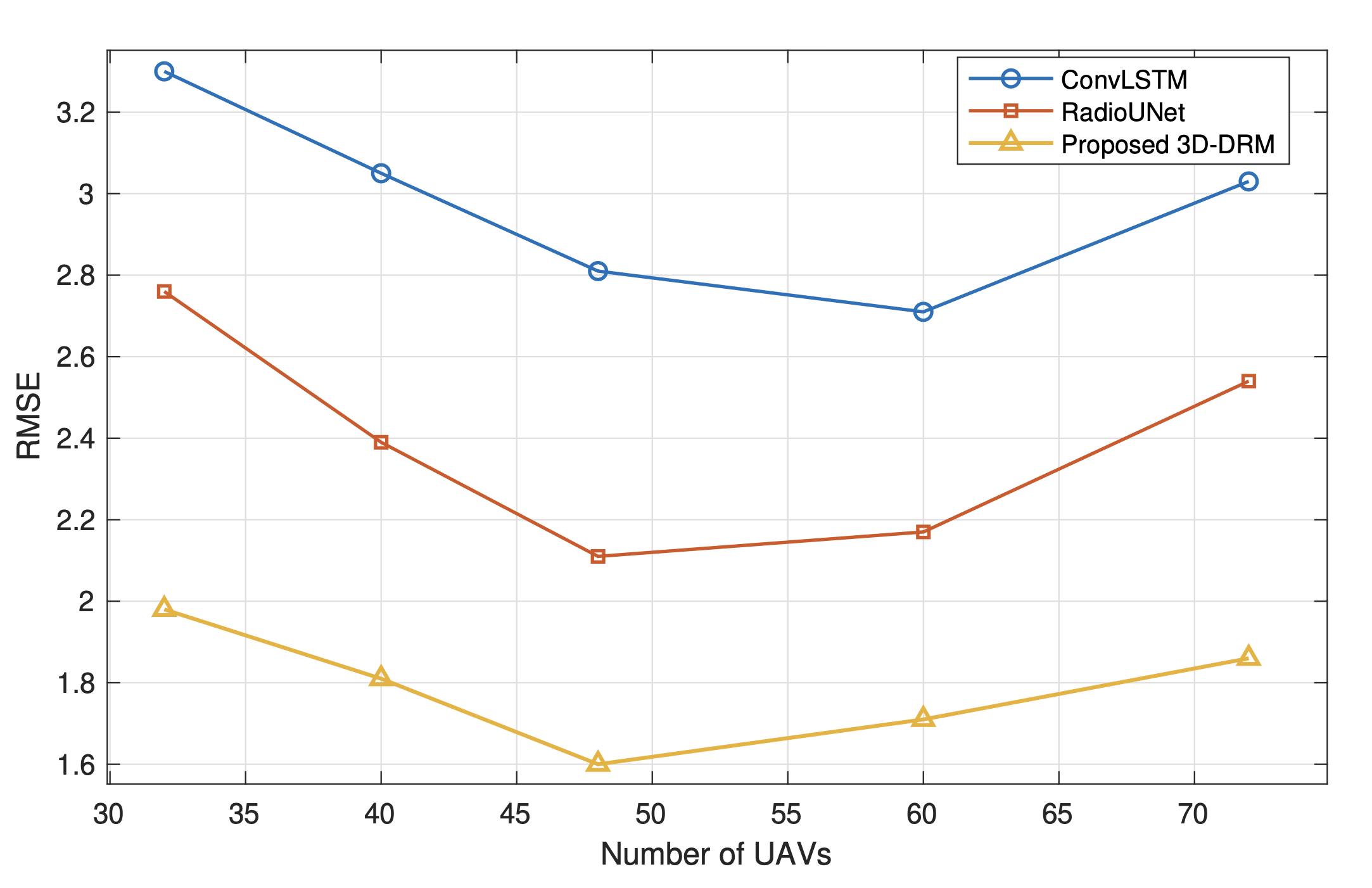}%
} &
\raisebox{-0.5\height}{%
    \includegraphics[width=0.35\textwidth]{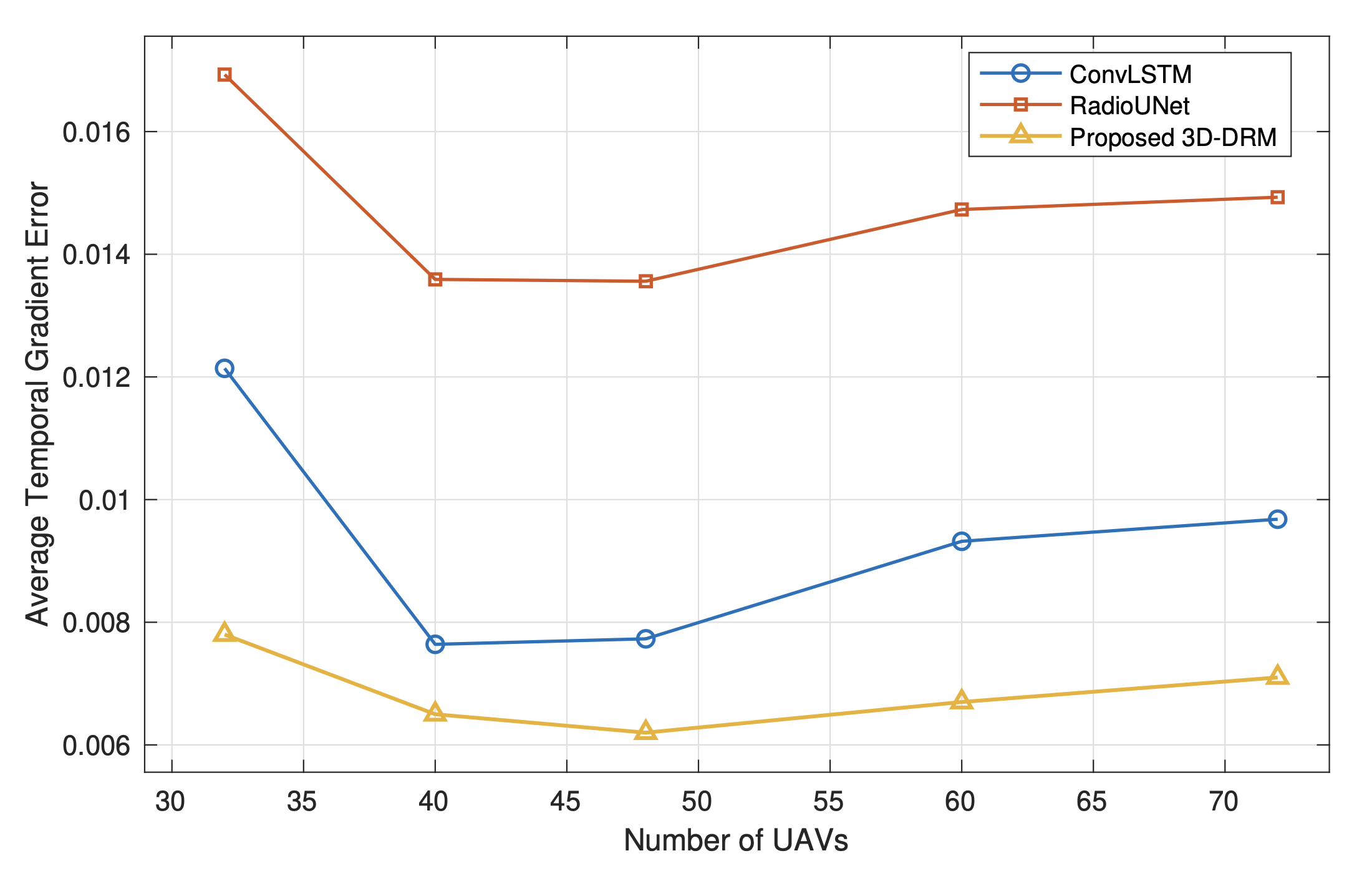}%
} &
\raisebox{-0.5\height}{%
    \includegraphics[width=0.35\textwidth]{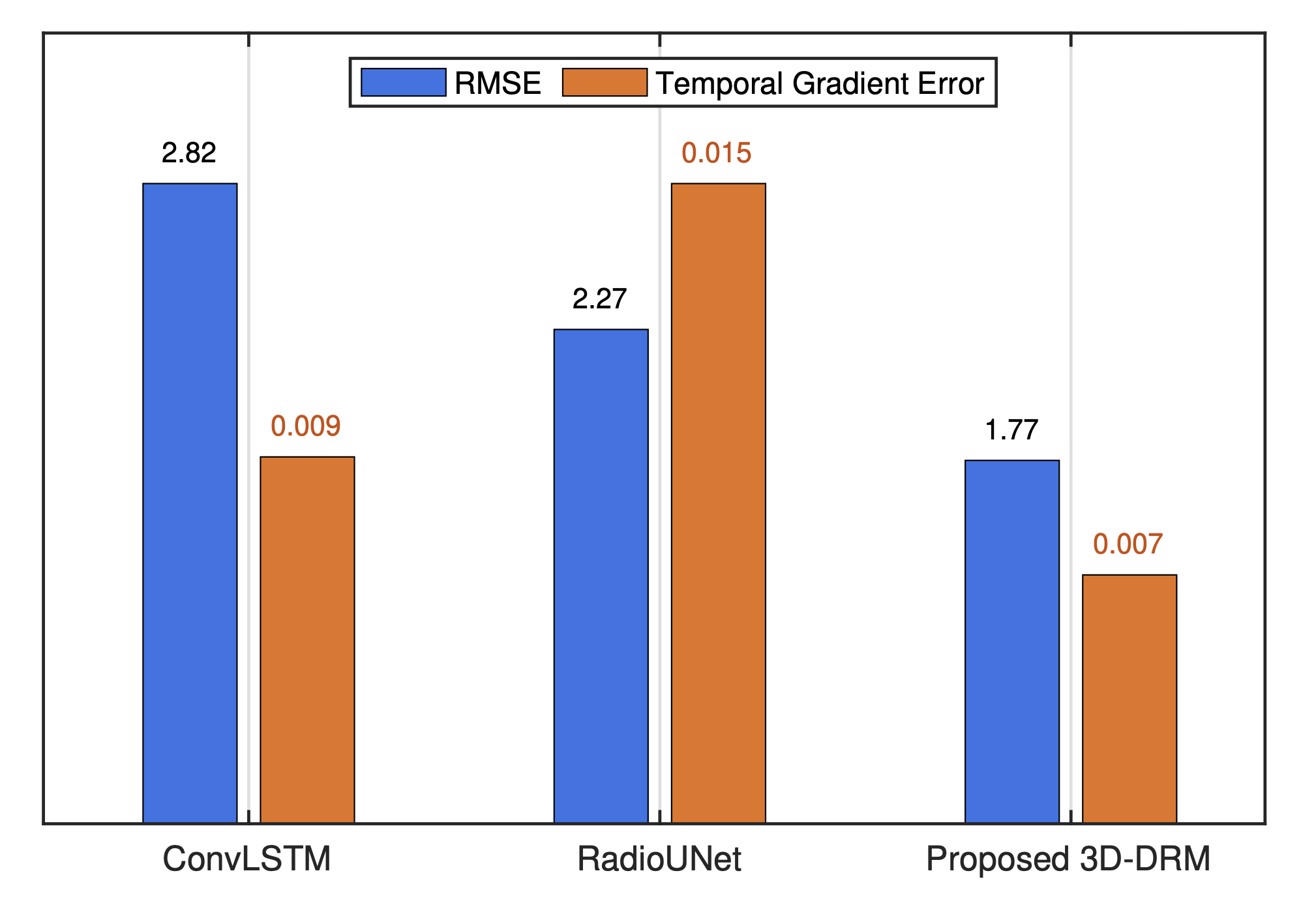}%
} \\
\small {(a)} &
\small {(b)} &
\small {(c)} \\
\end{tabular}

\caption{Quantitative comparison of model performance: 
(a) RMSE versus number of UAVs, 
(b) temporal gradient error versus number of UAVs, and 
(c) overall spatial and temporal performance comparison among ConvLSTM, RadioUNet, and the proposed 3D-DRM.}\vspace{-10pt}
\label{fig:eval_results}
\end{figure*}

\begin{figure}[t]
\centering
\vspace{-5pt}
\begin{tabular}{ c @{\hspace{5pt}} c }
\includegraphics[width=0.40\linewidth]
{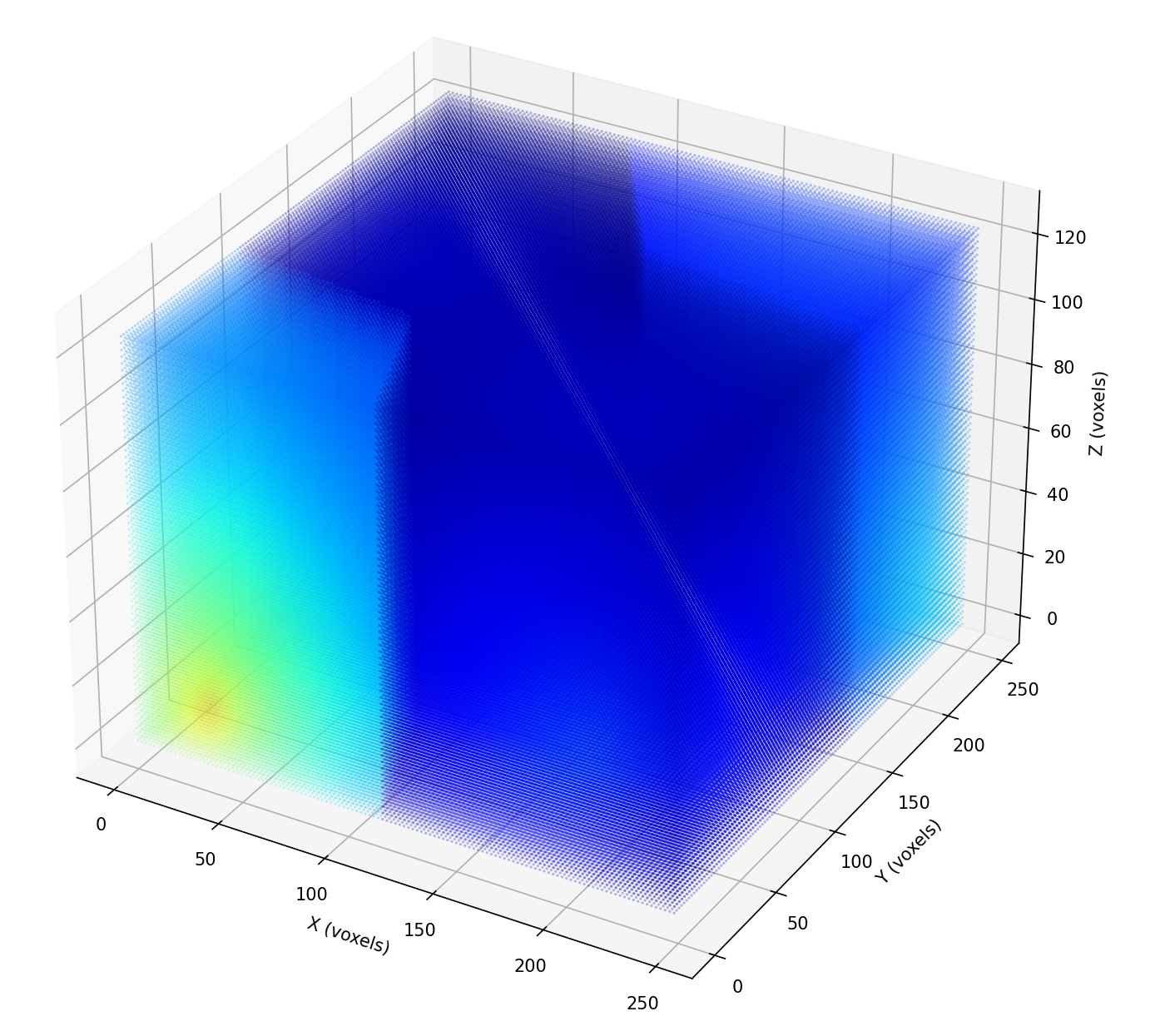} &
\includegraphics[width=0.48\linewidth]{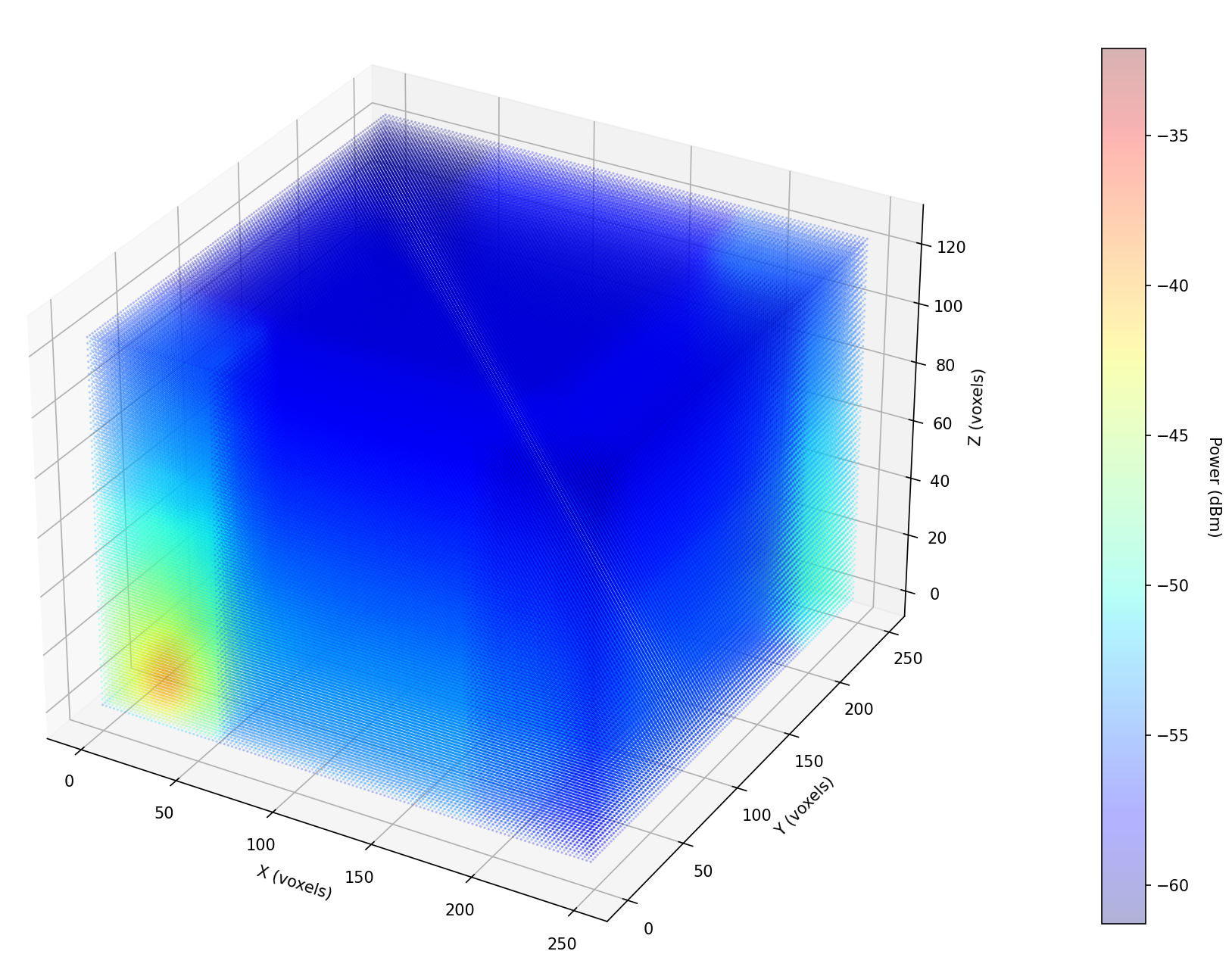} \\
\small (a) Ground-truth & \small (b) Predicted
\end{tabular}
\caption{Visualization of ground-truth and predicted 3D RMs.}
\label{fig:3d-map}
\end{figure}

The quantitative results in Fig.~\ref{fig:eval_results} demonstrate that the proposed 3D-DRM consistently outperforms both ConvLSTM and RadioUNet in terms of RMSE and temporal gradient error. As shown in Fig.~\ref{fig:eval_results}(a), 3D-DRM achieves the lowest RMSE across all UAV configurations, with an average reduction of approximately $35\%$ and $25\%$ compared to ConvLSTM and RadioUNet, respectively. Although RadioUNet is designed for static reconstruction, its purely convolutional design limits its ability to connect temporal dependencies and perform accurate forecasting, leading to noticeable degradation when predicting future maps. ConvLSTM, while tailored for temporal prediction, exhibits higher reconstruction errors due to limited spatial expressiveness. Moreover, as the number of UAVs increases, all models initially benefit from denser sampling, but the RMSE slightly rises beyond 60 UAVs, likely because a more dynamic propagation environment increases signal fluctuation and reconstruction difficulty.

Fig.~\ref{fig:eval_results}(b) further confirms the temporal advantage of the proposed model. Across four prediction steps, 3D-DRM maintains the lowest temporal gradient error, demonstrating smoother and more coherent temporal transitions than ConvLSTM and RadioUNet. This advantage arises from the ViT-based spatio-temporal attention mechanism, which effectively learns long-range dependencies in both spatial and temporal domains. In contrast, ConvLSTM captures short-term motion but struggles with long-horizon consistency, while RadioUNet lacks explicit temporal modeling. The joint comparison in Fig.~\ref{fig:eval_results}(c) highlights the superiority of 3D-DRM in both spatial and temporal metrics, which aligns with the qualitative visualization in Fig.~\ref{fig:3d-map}, where the predicted RMs exhibit sharper structures and smoother temporal evolution than the baselines.

From a deployment standpoint, 3D-DRM offers the most scalable solution, enabling parallel inference across spatial and temporal domains for real-time operation. In contrast, RadioUNet allows fast but non-sequential inference, while the recurrent design of ConvLSTM introduces latency due to step-by-step processing. Overall, 3D-DRM combines deployment efficiency with strong capability in reconstructing and forecasting 3D dynamic RMs. Future extensions could leverage techniques from spectrum sensing~\cite{liu2019deep, liu2019maximum, liu2017optimal} and cognitive radio~\cite{xie2020deep, xie2019activity} to enhance situational awareness.

\balance
\section{Conclusion}
This paper presented the 3D-DRM, a Transformer-based framework for predicting time-evolving 3D RMs in low-altitude wireless networks. By formulating the task as a unified spatio-temporal learning problem, 3D-DRM employs a ViT to encode spatial power distributions and a temporal Transformer to capture dynamic evolution over time. Experimental results under diverse network conditions demonstrate that the proposed method consistently outperforms deep learning baselines across a wide range of user densities and dynamic mobility patterns, exhibiting superior spatial accuracy and temporal consistency. Future work will extend this framework toward multi-layer RM prediction and distributed learning for real-time network optimization.

\bibliographystyle{IEEEtran}
\bibliography{refs}

\end{document}